\documentclass[letterpaper]{article} 
\usepackage{aaai2026}  
\usepackage{times}  
\usepackage{helvet}  
\usepackage{courier}  
\usepackage[hyphens]{url}  
\usepackage{graphicx} 
\usepackage{natbib}  
\usepackage{caption} 
\usepackage{algorithm}
\usepackage{algorithmic}
\usepackage{fontawesome}
\usepackage{enumitem}
\usepackage{multicol}
\usepackage{multirow}
\usepackage{xspace}
\usepackage{amssymb}
\usepackage{epigraph}
\usepackage{mathtools}
\usepackage{changepage}
\usepackage{booktabs}
\usepackage{xcolor}
\usepackage{makecell}

\usepackage{newfloat}
\usepackage{listings}
\DeclareCaptionStyle{ruled}{labelfont=normalfont,labelsep=colon,strut=off} 
\floatstyle{ruled}
\newfloat{listing}{tb}{lst}{}
\floatname{listing}{Listing}
\title{Stratos: An End-to-End Distillation Pipeline for Customized LLMs under Distributed Cloud Environments}
\author {
    Ziming Dai\textsuperscript{\rm 1}\thanks{Equal contribution.},
    Tuo Zhang\textsuperscript{\rm 2}\footnotemark[1],
    Fei Gao\textsuperscript{\rm 1}, 
    Xingyi Cai\textsuperscript{\rm 1}, 
    Xiaofei Wang\textsuperscript{\rm 1},
    Cheng Zhang\textsuperscript{\rm 3},
    Wenyu Wang\textsuperscript{\rm 4},
    Chengjie Zang\textsuperscript{\rm 4}
}
\affiliations {
    \textsuperscript{\rm 1}College of Intelligence and Computing, Tianjin University, Tianjin, China\\
    \textsuperscript{\rm 2}University of Southern California, Los Angeles, United States\\
    \textsuperscript{\rm 3}Institute of Technology, Tianjin University of Finance and Economics, Tianjin, China\\
    \textsuperscript{\rm 4}Paiou Cloud Computing (Shanghai) Co., Ltd, Shanghai, China\\
    phoenixdai@tju.edu.cn, tuozhang@usc.edu, g\_f@tju.edu.cn, xingyicai\_@outlook.com, xiaofeiwang@tju.edu.cn, zhangcheng@tjufe.edu.cn, wayne@pplabs.org, kaxifa@pplabs.org
}

\usepackage{bibentry}

\begin{document}

\maketitle

\begin{abstract}
The growing industrial demand for customized and cost-efficient large language models (LLMs) is fueled by the rise of vertical, domain-specific tasks and the need to optimize performance under constraints such as latency and budget. Knowledge distillation, as an efficient model compression and transfer technique, offers a feasible solution. However, existing distillation frameworks often require manual intervention and struggle to meet such complex user-defined distillation requirements. To bridge this gap, we propose \texttt{Stratos}, an end-to-end LLM distillation pipeline that automates server/model selection, knowledge distillation, and deployment in distributed cloud environments. Given user-defined constraints on model performance and system budget, \texttt{Stratos} automatically selects Pareto-optimal servers, dynamically matches teacher–student pairs, and adapts distillation strategies based on task complexity to optimize cloud hosting. Experiments show that \texttt{Stratos} produces a student model that achieves four times the accuracy of its GPT-4o teacher baseline on a rare, domain-specific Mahjong reasoning task with reverse synthetic data and knowledge injection. Moreover, it achieves reduced latency and cost without compromising accuracy. These results highlight its promise for vertical-domain LLM deployment.

\end{abstract}


\section{Introduction}

Large language models (LLMs), such as GPT-4o and Claude 3.5, have demonstrated impressive capabilities across diverse tasks.  However, their massive size, high computational cost, and limited controllability pose significant challenges for widespread deployment, especially in vertical domains where cost-efficiency, data privacy, or regulatory constraints are critical~\cite{Wan2023EfficientLL, brown2020language}. Many users, particularly in enterprise and industrial contexts, demand models that are not only accurate but also lightweight, customizable, and easy to host under constrained infrastructure.

\begin{figure}[t]
	\centering
   \includegraphics[width=\linewidth]{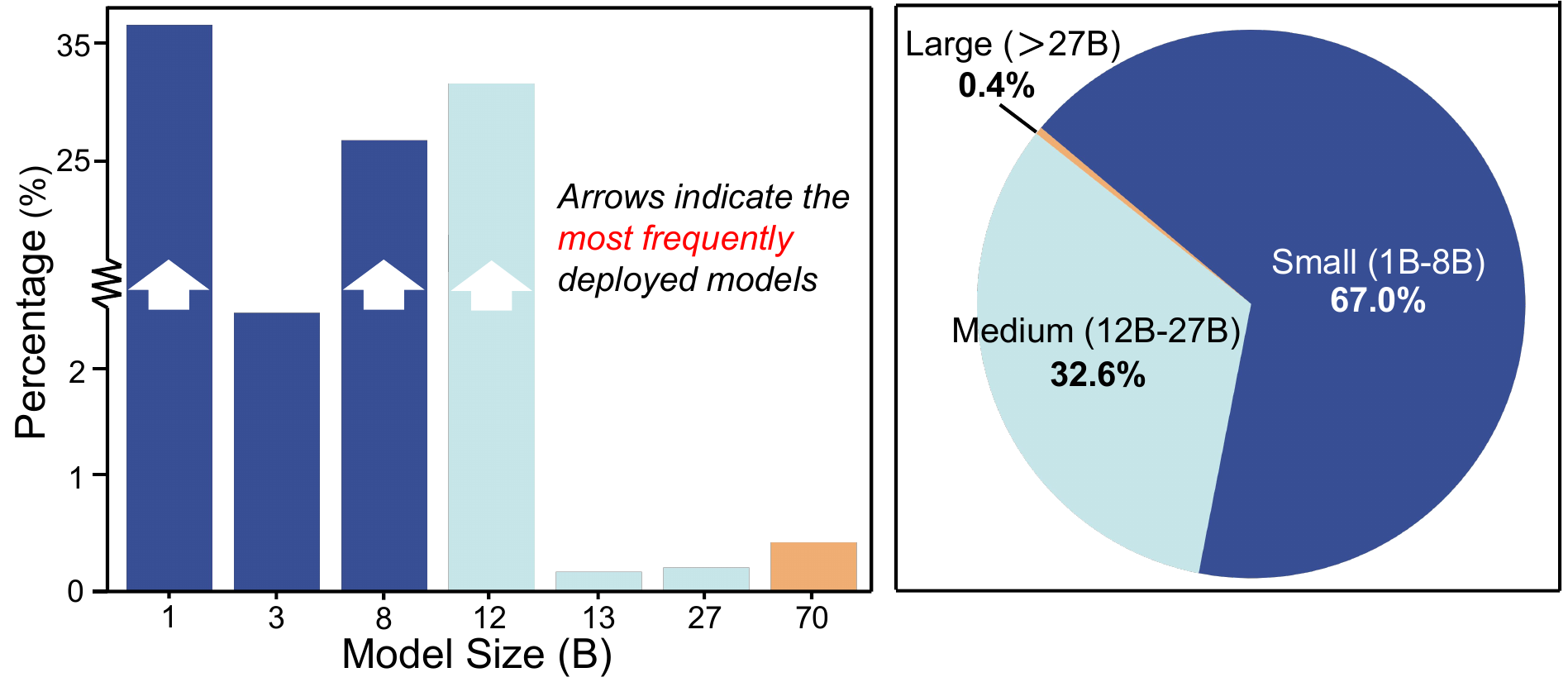}
	\caption{Industrial deployment distribution of LLMs by parameter size.}
	\label{fig:bg}
\end{figure}

The demand for customized and lightweight language models has moved beyond academic exploration and become a practical necessity in industrial deployments. As shown in Figure~\ref{fig:bg}, nearly a month of deployment statistics from Paiou Cloud Computing Corporation (PPIO) reveal that approximately 67\% of deployed models have fewer than 8 billion parameters. While frontier models such as Qwen-72B are occasionally utilized for zero-shot tasks, large-scale deployments predominantly rely on domain-specific small models. These observations reflect a growing industry consensus: smaller, controllable LLMs are essential for building scalable, cost-efficient, and robust AI infrastructure.

However, while knowledge distillation has been extensively studied, existing solutions remain fragmented and under-specified for deployment. Some focus solely on synthetic data generation~\cite{sky_t1_2025}, others on model fine-tuning~\cite{xu2024data}, and still others on compression or quantization~\cite{lin2024awq}. These techniques are often evaluated in isolation, assuming fixed model pairs and unconstrained training resources, thus ignoring real-world deployment constraints such as latency, training budget, or inference infrastructure. Few systems offer a unified solution that bridges task-specific model selection, strategy adaptation, and resource-aware deployment into a single pipeline.

\begin{figure}[t]
	\centering
   \includegraphics[width=\linewidth]{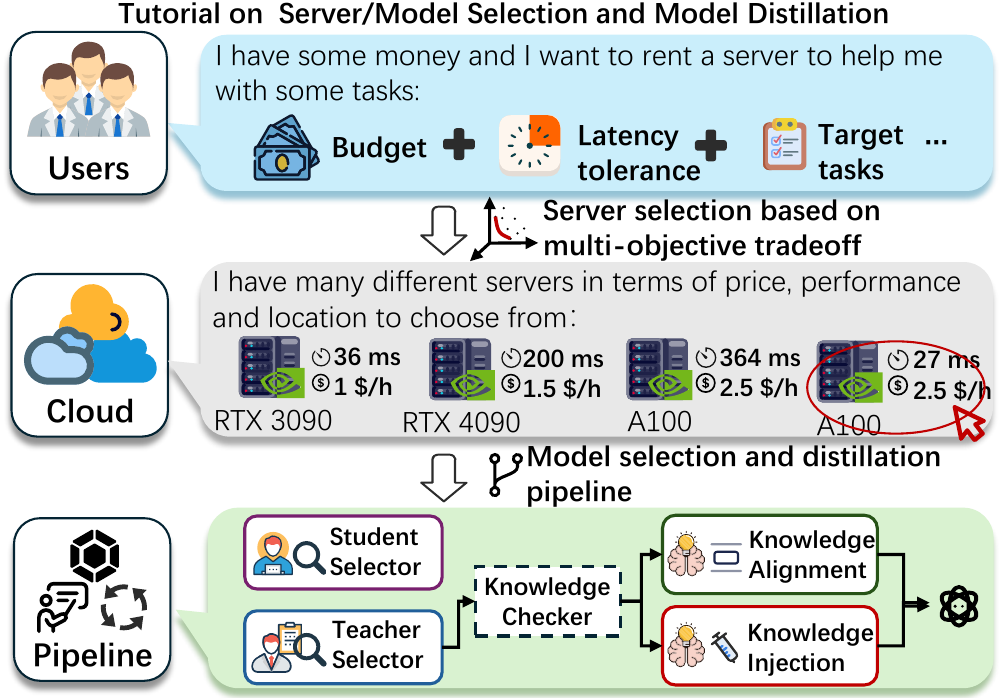}
	\caption{Real-world scenario and optimization entry point.}
	\label{fig:framework}
\end{figure}

Therefore, we propose \texttt{Stratos}, a unified pipeline for model distillation and deployment. Figure~\ref{fig:framework} illustrates the real-world scenario and optimization entry point, where user constraints are jointly optimized with heterogeneous cloud resources through an automated workflow. \texttt{Stratos} formulates the entire process as a meta-optimization problem that integrates: (i) Pareto frontier-guided server selection that balances latency, and hosting cost; (ii) dynamic teacher-student pairing based on task complexity and cost-performance trade-offs; and (iii) adaptive strategy selection between knowledge alignment and knowledge injection.

We have fully deployed and integrated \texttt{Stratos} into the distributed cloud environment of PPIO, a leading commercial platform for decentralized edge computing with over 3,000 servers worldwide. Known for pioneering large-scale AIGC deployment on crowdsourced infrastructure, it provides a highly dynamic and resource-constrained environment representative of real-world industrial challenges. We further evaluate it on a range of reasoning benchmarks, including GSM8K~\cite{cobbe2021gsm8k}, AIME, and a domain-specific Mahjong dataset. Across these tasks, \texttt{Stratos} produces student models that significantly outperform their teachers, achieving up to 4× accuracy gains in low-resource settings while respecting budget and latency constraints. 

In summary, \texttt{Stratos} is a cohesive and automated pipeline that lowers the barrier to building high-quality, domain-specific LLMs in production environments. Our contributions are as follows:
\begin{itemize}
    \item We present \texttt{Stratos}, the first end-to-end system to automate the entire lifecycle of LLM distillation and deployment. Unlike prior work that treats model selection, knowledge transfer, and server assignment separately, \texttt{Stratos} integrates them into a unified, task-aware pipeline that supports real-world constraints and scales to commercial deployment.
    
    \item We formulate deployment as a multi-objective optimization problem, where user-defined constraints—such as latency, and accuracy—are jointly optimized across heterogeneous cloud infrastructures. \texttt{Stratos} efficiently identifies Pareto-optimal teacher-student-server configurations using task-aware search and constraint modeling.
    
    \item We design a dual-mode knowledge distillation framework that adaptively switches between two strategies, based on the teacher’s capability for the target task. This enables \texttt{Stratos} to deliver strong performance even for tasks with low pretraining coverage, extending beyond the capabilities of frontier LLMs.
\end{itemize}

\section{Related Works} 
\textbf{LLM Distillation Strategies.} Recent research has explored various approaches to transferring knowledge from LLMs to smaller ones through distillation. Most methods rely on generating synthetic data for supervised fine-tuning~\cite{Zhang2023GPTFLGP}. Techniques like Impossible Distillation~\cite{jung2024impossible} identify subspace data to improve quality, while others~\cite{hsieh2023distilling, ramnath2023tailoring} perform distillation based on intermediate signals such as rationales. However, these works primarily aim to optimize training outcomes, and none address how to automatically select appropriate teacher-student pairs or how to incorporate deployment constraints into the distillation process—two core aspects tackled by \texttt{Stratos}.

\noindent \textbf{Synthetic Data Generation for Distillation.} Since the quality of synthetic data plays a crucial role in distillation effectiveness, recent studies have focused on novel methods for generating high-quality training data. Sky-T1 \cite{sky_t1_2025} selects datasets from multiple domains to create a diverse dataset and improves distillation efficiency by leveraging GPT-4o-mini to format the data after supplementing it with reasoning data from an LLM. OpenThought \cite{openthoughts} enhances data accuracy by verifying domain-specific reasoning data generated by DeepSeek-R1. Meanwhile, s1 \cite{muennighoff2025s1} applies data cleaning and filtering to multi-domain datasets, classifying data based on quality, diversity, and difficulty, ultimately obtaining 1,000 high-quality questions. In contrast, \texttt{Stratos} treats synthetic data generation as a sub-component of a larger decision pipeline, choosing teachers and prompting strategies adaptively based on domain coverage.

\noindent \textbf{System Frameworks for LLM Customization.} Recent frameworks like LLaMAFactory~\cite{zheng2024llamafactory} have supported zero-code fine-tuning of diverse LLMs, including methods like LoRA and QLoRA. However, they still rely on manual trial-and-error for teacher–student pairing and deployment choices. Additionally, AIBrix provides a cloud-native control plane tailored for scalable LLM inference~\cite{team2025aibrix}, focusing on deployment orchestration while overlooking the distillation process. In contrast, \texttt{Stratos} offers an end-to-end pipeline that automates the entire lifecycle of LLM distillation and deployment, driven by task requirements and user-defined constraints.

 \begin{figure*}[t]
    \centering
    \includegraphics[width=\textwidth]{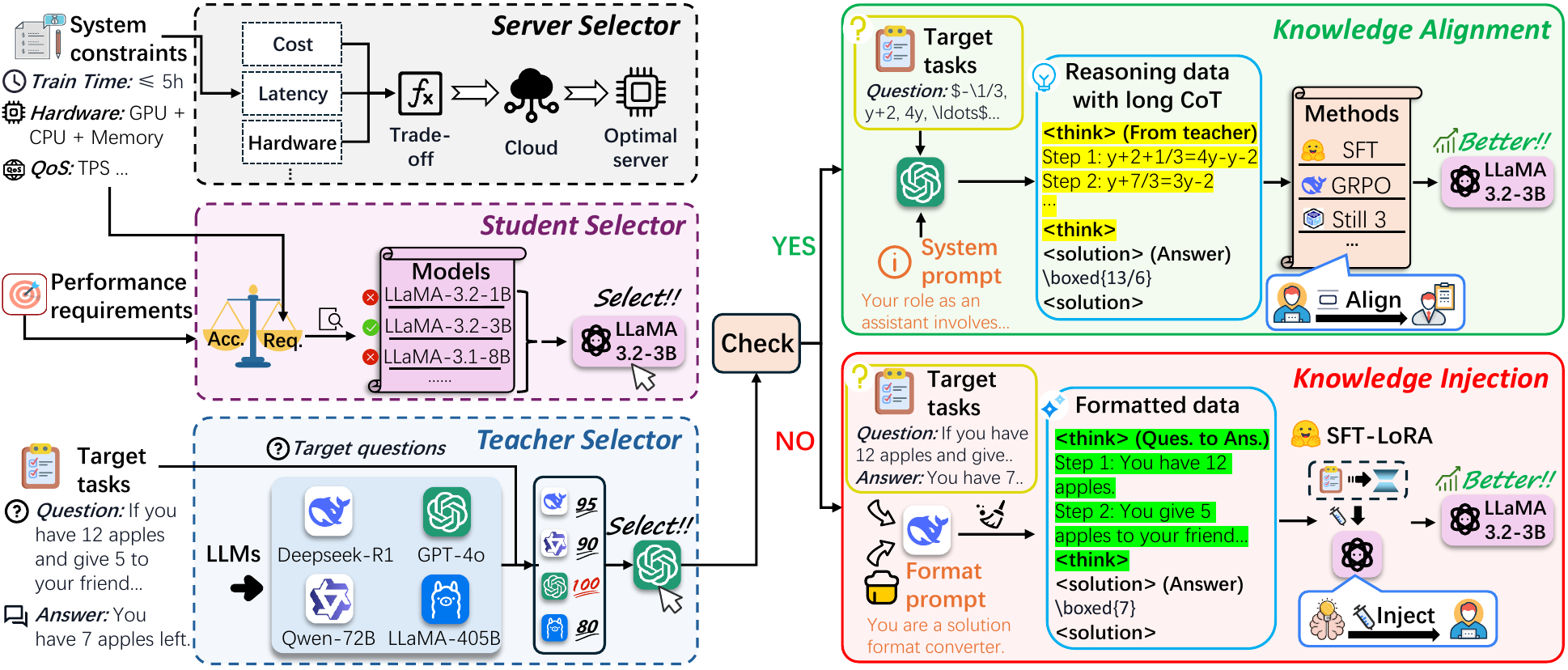}
    \caption{System overview of the proposed end-to-end distillation pipeline.}
    \label{fig:pipeline}
\end{figure*}

\section{Design Details} \label{sec:design}

\subsection{Overview of \texttt{Stratos}}
\texttt{Stratos} is designed as a fully automated pipeline that customizes and deploys expert LLMs under user-defined constraints. Given an input specification that includes a target task, performance requirements (e.g., minimum accuracy), and system constraints (e.g., cost budget, latency tolerance), \texttt{Stratos} performs a sequence of decisions across four interconnected components: Server Selector, Teacher Selector, Student Selector, and Distillation Strategy Selector. These components work together to optimize both model quality and system efficiency in distributed cloud environments.
    
As shown in Figure~\ref{fig:pipeline}, \texttt{Stratos} first selects a server with the appropriate hardware configuration from the cloud, balancing training and inference cost, hardware capability, and network latency. Based on the selected server and the target task, it then chooses an optimal teacher-student model pair, guided by task performance evaluation and system feasibility constraints. Depending on whether the selected teacher model demonstrates strong performance on the target task, \texttt{Stratos} dynamically selects one of two knowledge distillation strategies: Knowledge Alignment or Knowledge Injection. The former extracts structured reasoning traces from the teacher, while the latter supplements training data through guided teacher reasoning based on known answers.

The output is a customized student model, trained using the selected strategy, and deployed on the recommended server. This architecture enables \texttt{Stratos} to function as a turn-key LLM distillation pipeline that adapts to diverse task domains and resource budgets, reducing the need for manual intervention while improving performance-cost tradeoffs.

\subsection{Server Selector}
In distributed cloud environments, physical servers are available across different geographical locations, each varying in performance, cost, and accessibility. To ensure efficient deployment, the first step is identifying the optimal server that meets the user’s constraint, such as budget, latency tolerance, and computational requirements. \texttt{Stratos} automates this selection process, balancing trade-offs between multiple metrics to optimize training and hosting efficiency by leveraging the Pareto Front Grid (PFG) method~\cite{xu2023pareto}.

For users, an ideal server should meet the following criteria: (i) Sufficient hardware power to enable the model to achieve fast training and inference or other required performance benchmarks; (ii) low rental costs to minimize financial overhead; and (iii) low communication latency to ensure efficient interaction with the model. Thus, we need to balance these three factors when selecting a suitable server. We represent each cloud server offered by providers as a triplet $\mathcal{S}_i = (H_i, C_i, L_i)$, where $H_i$ denotes the hardware specifications of the server,  $C_i$ is the rental cost, and $L_i$ is the communication latency between the server and the user.

We introduce a window $\gamma$ to represent the acceptable range of rent cost fluctuations, and divide each objective into $K$ intervals, where $K = \lceil|C_* - C_-| / \gamma\rceil$. Here, $C_*$ and $C_-$ are the costs of the ideally best and worst servers.The entire objective space is partitioned into $K^3$ regions. Since hardware constraints are a mandatory requirement, we exclude servers with hardware specifications below the threshold from the objective space. We define the solution set $S^l(k)$ for the $l$-th objective within the $k$-th interval as:
\begin{equation}
 \begin{split}
    S^l(k)=\{\mathcal{S}_i|\Psi^1(\mathcal{S}_i)=1, \ldots,\Psi^{l-1}(\mathcal{S}_i)=l-1,\\
    \Psi^{l+1}(\mathcal{S}_i)=l+1,\ldots,\Psi^{K}(\mathcal{S}_i)=K
    \},
 \label{sub-problem_func}
 \end{split}
\end{equation}
where $\Psi^l(\mathcal{S}_i)$ represents the grid coordinate of server $\mathcal{S}_i$ on the $l$-th objective. For each objective function, we define $g_{l,k}^*$ as the optimal value among all solutions within the $k$-th interval $S_s^l(k)$ at grid coordinate $\Psi^l(\mathcal{S}_i)$. Then, we construct $\Phi_{l,k}$ as the set of solutions in $S^l(k)$ that satisfy $\Psi^l(\mathcal{S}_i) = g_{l,k}^*$. The union of all $\Phi_{l,k}$ forms the PFG. We select the grid with the lowest cost in the PFG and choose the server within it that has the shortest Euclidean distance to the ideal best server as the final selection~\cite{dai2025acme}. It is important to note that the PFG framework is independent of the exact goal and can be extended to other deployment environments by redefining scoring or constraint criteria.

\subsection{Teacher Selector}
Then, the system chooses a teacher model based on the user’s target task. With the emergence of powerful LLMs such as GPT-4o, Deepseek-R1~\cite{guo2025deepseek}, and Qwen-72B~\cite{qwen}, we have access to models with strong problem-solving capabilities. But due to differences in training data and methodologies, their performance varies across different task types. To identify the most suitable teacher model, \texttt{Stratos} evaluates a subset of the target tasks across multiple LLMs and computes their accuracy. If more than one model meets the user's accuracy requirement, we select the most cost-effective option for distillation. Otherwise, we prioritize the model with the highest accuracy to maximize knowledge transfer to the student model.

\subsection{Student Selector}
Considering the diversity of user tasks and the selected servers, the selection of available models must be adjusted accordingly. User demands can be abstracted into two components: an accuracy threshold $A$, representing the minimal acceptable performance for a given task, and a set of system requirements $\mathbb{R}$, including service quality requirements ($R$: token per second (TPS)), allowable training time ($T$). These parameters jointly define the feasible search space for candidate student models.

\begin{table}[t]
    \centering
    \small
    \begin{tabular}{lcc}
    \toprule
    \textbf{Model} & \textbf{Acc. (No 8B CoT)} & \textbf{Acc. (With 8B CoT)} \\
    \midrule
    LLaMA-3.1-8B     & 61.49\%               & -- \\
    LLaMA-3.2-1B     & 40.79\%               & 44.20\% (3.41\%\textuparrow) \\
    Qwen-2.5-1.5B    & 55.34\%               & 57.32\% (1.98\%\textuparrow) \\
    LLaMA-3.2-3B     & 30.02\%               & \textbf{59.67\% (29.65\%\textuparrow)} \\
    \bottomrule
    \end{tabular}
    \caption{Impact of injecting LLaMA-3.1-8B reasoning steps into smaller models on GSM8K.}
    \label{tab:llama_gsm8k}
\end{table}

However, not all candidate models are equally capable of learning from teacher models through distillation. In particular, smaller models may lack the capacity to process structured reasoning, rendering the distillation ineffective. To investigate this, we conducted an ablation experiment where structured reasoning steps from LLaMA-3.1-8B were used as prompts for the GSM8K dataset. As shown in Table~\ref{tab:llama_gsm8k}, while LLaMA-3.2-3B showed a marked improvement over its baseline, LLaMA-3.2-1B and Qwen-2.5-1.5B yielded negligible gains. This indicates that below a certain parameter threshold, models cannot benefit from reasoning-rich supervision, and should thus be excluded from candidate selection. Moreover, hardware limits impose upper bounds, which are based on GPU memory and runtime cost.

To determine which models can meet the accuracy demands for the given task, we sample the target task and use the selected teacher model to generate reasoning steps, which are then provided as
prompts to student models. If a model's accuracy $a$ does not meet the required threshold $A$, it is discarded. For the remaining models, we apply a weighted scoring method to evaluate their suitability:
\begin{equation}
\begin{split}
    M = \alpha \cdot \tilde{a} + \beta \cdot (1 - \tilde{t}) + \theta \cdot \tilde{r}, \quad \text{s.t.} \ t \leq T,\ r \geq R,
 \label{score_func}
 \end{split}
\end{equation}
where $t$ denotes the expected training time of the model on the server, $r$ is the expected TPS. The terms $\tilde{t}$, $\tilde{r}$, and $\tilde{a}$ refer to the normalized values of training time, throughput, and accuracy, respectively. The parameters $\alpha$, $\beta$, and $\theta$ are user-defined weights. Finally, we select the model with the highest score as the student model.

\subsection{Knowledge Alignment}
When the target task falls within the knowledge space of the teacher model, \texttt{Stratos} adopts a knowledge alignment strategy to reliably extract and transfer reasoning patterns. This strategy is automatically triggered when the teacher model demonstrates consistently high accuracy on the samples. As shown by the empirical results in Table~\ref{tab:llama_gsm8k}, we observe that the reasoning capability of the teacher directly affects the student model’s performance. Based on this insight, \texttt{Stratos} extracts and distills structured reasoning steps from the teacher model in the form of chain-of-thought, enabling the student model not only to generate correct answers but also to internalize advanced reasoning strategies. Given a target task, we prompt the teacher to explicitly articulate its thought process, converting raw responses into structured reasoning data.

The alignment module is designed to be plug-and-play and task-adaptive. In this paper, we provide two instantiations: (i) LoRA-based Supervised Fine-Tuning (SFT)~\cite{zheng2024llamafactory}, and (ii) Group Relative Policy Optimization (GRPO)~\cite{guo2025deepseek}. Users can easily swap in alternative distillation algorithms depending on hardware and latency constraints. This modular design allows \texttt{Stratos} to support scalable, low-cost, and controllable deployment of customized LLMs across tasks.

\subsection{Knowledge Injection}
When the target task falls outside the teacher model’s knowledge domain—indicated by low accuracy or inability to generate reliable reasoning traces—\texttt{Stratos} switches to a knowledge injection mode. In this setting, the teacher lacks sufficient pretrained exposure to directly solve or explain the task via prompting. So, we provide both the question and correct answer as input to the teacher and instruct it to generate plausible reasoning paths that bridge the two. This reverse reasoning process enables data synthesis even when forward inference fails. To enrich task coverage and support low-resource domains, \texttt{Stratos} uses real data as seeds to generate synthetic \texttt{Stratos} via structured prompting. To ensure data quality, we implement a rejection sampling mechanism using an independent teacher model instance as a verifier. Only samples where the reasoning and final answer are independently validated are retained, ensuring consistency and logical soundness.

Due to the teacher model’s lack of prior knowledge in the target domain, reinforcement learning (RL) and similar methods are unsuitable for this scenario as they do not provide any external knowledge injection. Instead, we employ LoRA-based SFT to inject knowledge into the student model. The SFT process is conducted on the reformatted dataset and synthetic data, effectively combining the teacher model’s reasoning capability with domain-specific knowledge from real data. This strategy allows \texttt{Stratos} to operate across tasks with varying domain familiarity, ensuring coverage and robustness in industrial deployment.

\begin{table*}[t]

    \centering
    \small
    \begin{tabular}{l c c c c c c c c c c}
        \toprule
        \textbf{Student Model} & \textbf{GRPO} & \textbf{SFT-LoRA} & \textbf{GSM8K} & \textbf{AIME 2022} & \textbf{AIME 2023} & \textbf{AIME 2024} & \textbf{Decision Reason} \\
        \midrule
        \multirow{3}{*}{\textcolor{red}{\faTimesCircle} LLaMA-3.2-1B}  
                      & \textcolor{red}{$\times$} & \textcolor{red}{$\times$} & 40.79\% & 23.33\% & 23.33\% & 20.00\%   
                      & \multirow{3}{*}{\makecell[l]{Limited learning capacity,\\accuracy below threshold.}}\\
                      & \checkmark & \textcolor{red}{$\times$} & 45.79\% & 36.67\% & 33.33\% & 40.00\% \\
                      & \textcolor{red}{$\times$} & \checkmark & 35.48\% & 46.67\% & 40.00\% & 53.33\% \\
        \cmidrule(lr){1-8}
        \multirow{3}{*}{\textcolor{red}{\faTimesCircle} Qwen-2.5-1.5B}  
                      & \textcolor{red}{$\times$} & \textcolor{red}{$\times$} & 55.34\% & 30.00\% & 30.00\% & 36.67\%   
                      & \multirow{3}{*}{\makecell[l]{Limited learning capacity,\\accuracy below threshold.}}\\
                      & \checkmark & \textcolor{red}{$\times$} & 61.39\% & 43.33\% & 33.33\% &40.00\%  \\
                      & \textcolor{red}{$\times$} & \checkmark & 64.67\% & 46.67\% &46.67\%& 53.33\% \\
        \cmidrule(lr){1-8}
        \multirow{3}{*}{\textcolor{green}{\faCheckCircle} LLaMA-3.2-3B}  
                      & \textcolor{red}{$\times$} & \textcolor{red}{$\times$} & 30.02\% & 50.00\% & 56.67\% & 50.00\%  
                      & \multirow{3}{*}{\makecell[l]{Balanced performance \\and deployability.}}\\
                      & \checkmark & \textcolor{red}{$\times$} & 65.81\% & 53.33\% & 60.00\% & 66.67\% \\
                      & \textcolor{red}{$\times$} & \checkmark & 71.95\% & 66.67\% & 66.67\% & 53.33\% \\
        \cmidrule(lr){1-8}
        \multirow{3}{*}{\textcolor{red}{\faTimesCircle} LLaMA-3.1-8B}  
                      & \textcolor{red}{$\times$} & \textcolor{red}{$\times$} & 61.49\% & 30.00\% & 36.67\% & 46.67\% 
                      & \multirow{3}{*}{\makecell[l]{Exceeds selected server’s \\ hardware limits.}}\\
                      & \checkmark & \textcolor{red}{$\times$} & 61.94\% & 56.67\% & 50.00\% & 53.33\% \\
                      & \textcolor{red}{$\times$} & \checkmark & 78.77\% & 50.00\% & 46.67\% & 46.67\% \\
        \bottomrule
    \end{tabular}
    \caption{Performance comparison of LLaMA models across different benchmarks. LLaMA-3.2-1B and Qwen-2.5-1.5B are discarded due to their accuracy not meeting the threshold, while LLaMA-3.1-8B is excluded because of its excessive parameter size. Ultimately, LLaMA-3.1-3B is selected as the student model.}
    \label{tab:llama_performance}

\end{table*}

\section{Empirical Study}
\subsection{Experimental Setup}
\textbf{Distributed Cloud Environment Setup.} We deployed the \texttt{Stratos} system in PPIO's cloud environment, which  consists of 3,732 servers distributed across multiple locations in Asia. These servers feature a diverse set of GPU configurations, including RTX 4090, H20, RTX 6000 Ada, RTX 3090, L20, and A100 XSM4.

\noindent \textbf{LLM Selection.} For the teacher model pool, we include GPT-4o, Qwen-72B, and LLaMA-3.1-405B. For the student model pool, we select LLaMA-3.2-1B, Qwen-2.5-1.5B, LLaMA-3.2-3B, and LLaMA-3.1-8B.

\noindent \textbf{Tasks and Datasets.} We evaluate \texttt{Stratos} on two distinct task types, categorized based on whether the teacher model possesses the necessary domain knowledge. For tasks within the teacher model's knowledge space, we select mathematical reasoning and use the GSM8K and AIME datasets for evaluation. For the AIME task, due to insufficient accuracy of student models, we augmented prompts with 50\% solutions extracted from each sample to enable effective evaluation. The distilled student model is required to achieve an accuracy above 65\% on GSM8K.
\begin{figure}[t]
    \centering
    \includegraphics[width=\linewidth]{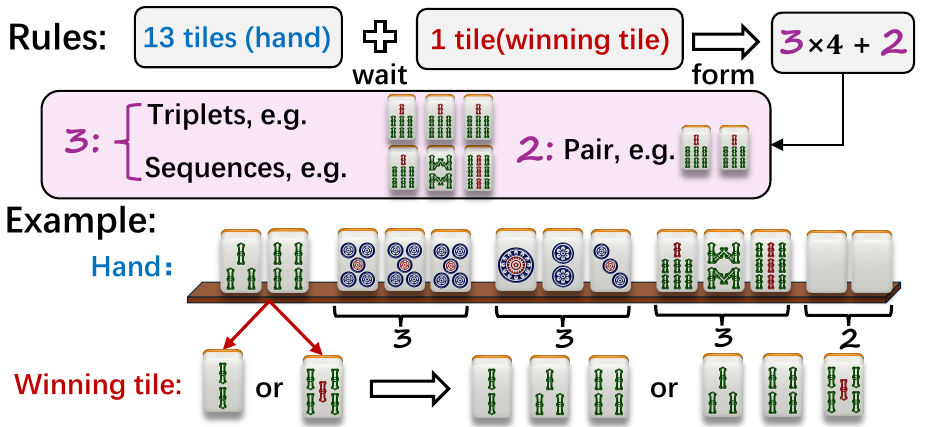}
    \caption{Illustration of Mahjong winning rules.}
    \label{fig:rule}

\end{figure}
\begin{table}[t]
    \centering
    \small
    \begin{tabular}{l c c c}
        \toprule
        \textbf{Model} & \textbf{GPT-4o} & \textbf{Claude 3.5 Sonnet} & \textbf{DeepSeek-R1} \\
        \midrule
        \textbf{Accuracy}       & 4\%  & 9\%  & 21\% \\
        \bottomrule
    \end{tabular}
    \caption{Performance of teacher models on the Mahjong-Winning-Tiles dataset.}
    \label{tab:teacher_performance}
\end{table}

For tasks beyond the teacher model's knowledge space, we select a Mahjong-related reasoning task, a rare and complex domain. It is a classic four-player tile-based game that originated in China. Figure~\ref{fig:rule} shows the game's rules and examples. The objective of Mahjong is to form a complete hand with 13 tiles, plus one additional tile, by creating four melds and one pair. A meld can be either a sequence (three consecutive tiles of the same suit) or a triplet (three identical tiles), while a pair consists of two identical tiles.

For evaluation, we use the Mahjong-Winning-Tiles dataset~\cite{mahjong-winning-tiles}, which contains 4,260 training entries and 100 test entries, each consisting of a hand and the corresponding winning tile. We input the models with the hand tile and ask for the required winning tile. Since each tile in this dataset are represented in its own unicode, existing LLMs have never encountered similar knowledge during pretraining. Moreover, Mahjong requires pattern recognition, probability estimation, and complex tile combinations, making it a challenging benchmark for models lacking prior exposure to the domain. As shown in Table~\ref{tab:teacher_performance}, existing LLMs struggle significantly with this task. The distilled model is required to achieve an accuracy above 15\%.

\subsection{Strategy Effectiveness across Multi-Objective Metrics}

To assess the effectiveness of our strategy selection mechanism, we compare \texttt{Stratos} against three baselines: accuracy-first, cost-first, and random selection. As shown in Table~\ref{tab:strategy_comparison}, \texttt{Stratos} achieves the best overall score by balancing accuracy, latency, cost, and training time.  The overall score is computed as a weighted average of these four metrics, reflecting the trade-offs faced in real-world deployments.  While the accuracy-first approach yields the highest task performance, it incurs substantial latency and training overhead.  In contrast, \texttt{Stratos} maintains competitive accuracy while significantly reducing latency and training time, improving its overall score by at least 22\%, demonstrating its advantage in resource-constrained environments.

\begin{table}[t]
    \centering
    \small
    \begin{tabular}{lccccc}
        \toprule
        Strategy & Acc. & Latency & Cost & Time & Score \\
        \midrule
        \textbf{\texttt{Stratos}} & 59.24\% & 20.29 & 24 & 1.59 & \textbf{0.55} \\
        Acc.-First & 80.28\% & 45.44 & 645 & 40.42 & 0.25 \\
        Cost-First     & 42.36\% & 90.73 & 35 & 2.55 & 0.37 \\
        Random          & 62.84\% & 118.53 & 37 & 2.44 & 0.34 \\
        \bottomrule
    \end{tabular}
    \caption{Comparison of different strategy selection methods across four criteria and an aggregated overall score.}
    \label{tab:strategy_comparison}
\end{table}
\subsection{Distillation Performance within Teacher's Knowledge Domain}

We first evaluate the distillation performance within the teacher model's knowledge domain, using mathematical reasoning tasks. All models in the teacher model pool surpass the accuracy threshold, and Qwen-72B is selected as the teacher model due to its lowest inference cost. With the student model selector, we found the 1B and 1.5B model are not capable of the requirement, as indicated in Table~\ref{tab:llama_gsm8k}. The 3B model is selected by the optimizer. During the distillation process, we prompt the teacher model to inject structured reasoning steps into the training data from GSM8K and generate additional synthetic data based on real examples, expanding the distillation dataset to 10,000 samples. We leverage the OpenThought package~\cite{openthoughts} for synthetic data curation. To assess the generalization capability of the distilled model, we evaluate it on both GSM8K and AIME 2022–2024, as AIME represents a distinct mathematical reasoning dataset not directly used during distillation.

As shown in Table~\ref{tab:llama_performance}, the results demonstrate a significant improvement in the student model with distillation compared to training alone. However, the 1B and 1.5B models fail to reach the target accuracy, even after distillation, reinforcing our earlier observation that extremely small models lack the capacity to effectively leverage reasoning and align with the Student Selector’s predictions. We also observe that RL-based distillation methods are less stable compared to SFT-based approaches, which consistently yield higher performance improvements. This instability may arise from the difficulty in optimizing reinforcement learning signals for reasoning-heavy tasks.
Distillation not only improves performance on GSM8K but also enhances generalization to unseen mathematical reasoning tasks, as evidenced by accuracy gains on AIME 2022–2024. This suggests that distillation enhances reasoning skills, enabling the student model to generalize to novel tasks.

\subsection{Distillation Performance outside Teacher's Knowledge Domain}

In this section, we evaluate scenarios where teacher models lack domain-specific knowledge using the Mahjong dataset. As shown in Table~\ref{tab:teacher_performance}, current LLMs perform poorly, with even DeepSeek-R1 achieving only 21\%. Therefore, unlike the previous section, we cannot rely on prompting the teacher with only the question to construct reasoning steps.

\begin{figure}[t]
	\centering
   \includegraphics[width=\linewidth]{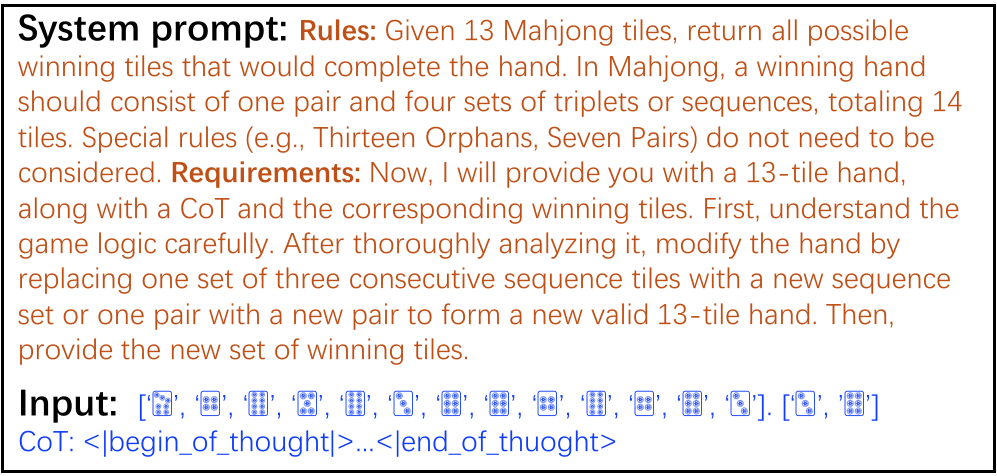}
	\caption{Example of synthetic Mahjong data generation.}
	\label{fig:prompt}

\end{figure}

\begin{table}[t]
    \centering
    \small
    \begin{tabular}{l c c c}
        \toprule
        \textbf{Model} & \textbf{0-shot} & \textbf{Original} & \textbf{Curation} \\
        \midrule
        LLaMA-3.2-1B       & 0.00\%  & 1.28\%  & \textbf{7.05\%} \\
        Qwen-2.5-1.5B       & 1.92\%  & 7.05\%  & \textbf{10.26\%} \\
        LLaMA-3.2-3B       & 4.49\%  & 7.69\%  & \textbf{17.31\%} \\
        LLaMA-3.1-8B       & 3.21\%  & 8.97\%  & \textbf{16.67\%} \\
        \bottomrule
    \end{tabular}
    \caption{Accuracy comparison of different models with original data and curation data.}
    \label{tab:model_performance}
\end{table}

To bridge this knowledge gap, we adopt a knowledge injection strategy by combining structured reasoning with synthetic data generation. Specifically, we prompt the teacher model with a hand and the corresponding winning tile to infer the intermediate reasoning steps required for making correct decisions, and then use these to train the student model. However, as shown in the “Original” row of Table~\ref{tab:model_performance}, although the original dataset can significantly improve the student model’s performance, it is still insufficient for fully transferring the teacher’s knowledge to a smaller model.

To further augment the training data, we generate an additional 4,000 synthetic samples using the teacher model, following the prompt structure illustrated in Figure~\ref{fig:prompt}. These synthetic examples are then combined with the original dataset to improve the student’s ability to generate reasoning steps during training.

Empowered by both structured reasoning and synthetic data, the student model fine-tuned via LoRA-based SFT achieves a 4× accuracy improvement over GPT-4o. This result confirms that combining the teacher model’s reasoning capacity with domain-specific insights enables the student model to acquire knowledge beyond the original limitations of the teacher, significantly improving both accuracy and reasoning performance.

\section{Conclusion}

We present \texttt{Stratos}, a deployment-oriented LLM distillation pipeline that automates server/model selection and distillation strategies in distributed cloud environments. \texttt{Stratos} improves performance while balancing accuracy and cost, showing strong practicality and broad applicability. 
The system is designed to serve as a reusable toolkit for LLM customization in real-world environments.
\bibliography{aaai2026}
\end{document}